\documentclass[letterpaper, 10 pt, conference]{ieeeconf}  %

\IEEEoverridecommandlockouts                              %

\usepackage{microtype}
\usepackage{graphicx}
\usepackage[colorlinks=True, citecolor=blue, linkcolor=.]{hyperref}
\usepackage[utf8]{inputenc} %
\usepackage[T1]{fontenc}    %
\usepackage{url}            %
\usepackage{booktabs}       %
\usepackage{amsfonts}       %
\usepackage{color}
\usepackage{mathtools}
\usepackage{amsmath,amssymb}
\usepackage[svgnames]{xcolor}
\usepackage{bm}
\usepackage{algorithm}
\usepackage{algorithmic}
\usepackage{subcaption}
\usepackage{hyperref}

\newcommand{\ie}{i.e., }
\newcommand{\eg}{e.g., }
\newcommand{\Skip}[1]{}
\renewcommand{\algorithmiccomment}[1]{\hfill\footnotesize
{\textcolor{gray}{$\triangleright$ #1}}}

\usepackage{array}
\newcommand{\PreserveBackslash}[1]{\let\temp=\\#1\let\\=\temp}
\newcolumntype{C}[1]{>{\PreserveBackslash\centering}p{#1}}
\newcolumntype{R}[1]{>{\PreserveBackslash\raggedleft}p{#1}}
\newcolumntype{L}[1]{>{\PreserveBackslash\raggedright}p{#1}}

\definecolor{red}{RGB}{230, 120, 120}
\definecolor{blue}{rgb}{0.4, 0.6, 0.8}
\definecolor{yellow}{rgb}{1.0, 0.88, 0.21}

\usepackage{amsmath}

\usepackage{amsmath,amsfonts,bm}

\def\eqref#1{equation~\ref{#1}}

\def\1{\bm{1}}

\DeclareMathAlphabet{\mathsfit}{\encodingdefault}{\sfdefault}{m}{sl}
\SetMathAlphabet{\mathsfit}{bold}{\encodingdefault}{\sfdefault}{bx}{n}

\DeclarePairedDelimiterX{\norm}[1]{\lVert}{\rVert}{#1}

\title{\LARGE \bf
Leveraging Scene Embeddings \\ for Gradient-Based Motion Planning in Latent Space
\author{Jun Yamada$^{\ast 1}$, %
Chia-Man Hung$^{\ast 1, 2}$,
Jack Collins$^{1}$,
Ioannis Havoutis$^{2}$,
Ingmar Posner$^{1}$ %
}

\thanks{$^{\ast}$Equal contribution.}
\thanks{$^{1}$Applied AI Lab (A2I), $^{2}$Dynamic Robot Systems (DRS)}%
\thanks{Oxford Robotics Institute (ORI), University of Oxford}
\thanks{Correspondence to: {\tt\small jyamada@robots.ox.ac.uk}}%
\thanks{Project page: \url{https://amp-ls.github.io/}}%
}

\renewcommand{\baselinestretch}{0.93}
\begin{document}

\maketitle
\thispagestyle{empty}
\pagestyle{empty}

\setlength{\abovedisplayskip}{6pt}
\setlength{\belowdisplayskip}{6pt}

\begin{abstract}

Motion planning framed as optimisation in structured latent spaces has recently emerged as competitive with traditional methods in terms of planning success while significantly outperforming them in terms of computational speed. However, the real-world applicability of recent work in this domain remains limited by the need to express obstacle information directly in \emph{state-space}, involving simple geometric primitives. In this work we address this challenge by leveraging learned scene embeddings together with a generative model of the robot manipulator to drive the optimisation process. In addition, we introduce an approach for efficient collision checking which directly regularises the optimisation undertaken for planning. Using simulated as well as real-world experiments, we demonstrate that our approach, AMP-LS, is able to successfully plan in novel, complex scenes while outperforming traditional planning baselines in terms of computation speed by an order of magnitude. We show that the resulting system is fast enough to enable closed-loop planning in real-world dynamic scenes.

\end{abstract}
\section{Introduction}
Motion planning is a core capability for robotic manipulation tasks~\cite{yamada2020mopa, xia2020relmogen} with the fundamental aim of planning a collision-free path from the current state of an articulated configuration of joints to a predefined goal joint or end-effector pose configuration. Sampling-based motion planning algorithms, such as Rapidly-Exploring Random Trees (RRT)~\cite{lavalle1998RRT} and Probabilistic Roadmap (PRM)~\cite{amato1996PRM}, are widely used within the robotics community as they have well understood properties in regards to planning time and collision avoidance. However, sampling-based methods become increasingly intractable as the problem size increases (\ie Degrees-of-Freedom (DoF) of the robot, environment complexity, and length of the path) and are also typically too slow to be used for closed-loop planning, as any change to the environment requires re-planning~\cite{short2016recent}. 

\begin{figure}
    \centering
    \includegraphics[width=\linewidth]{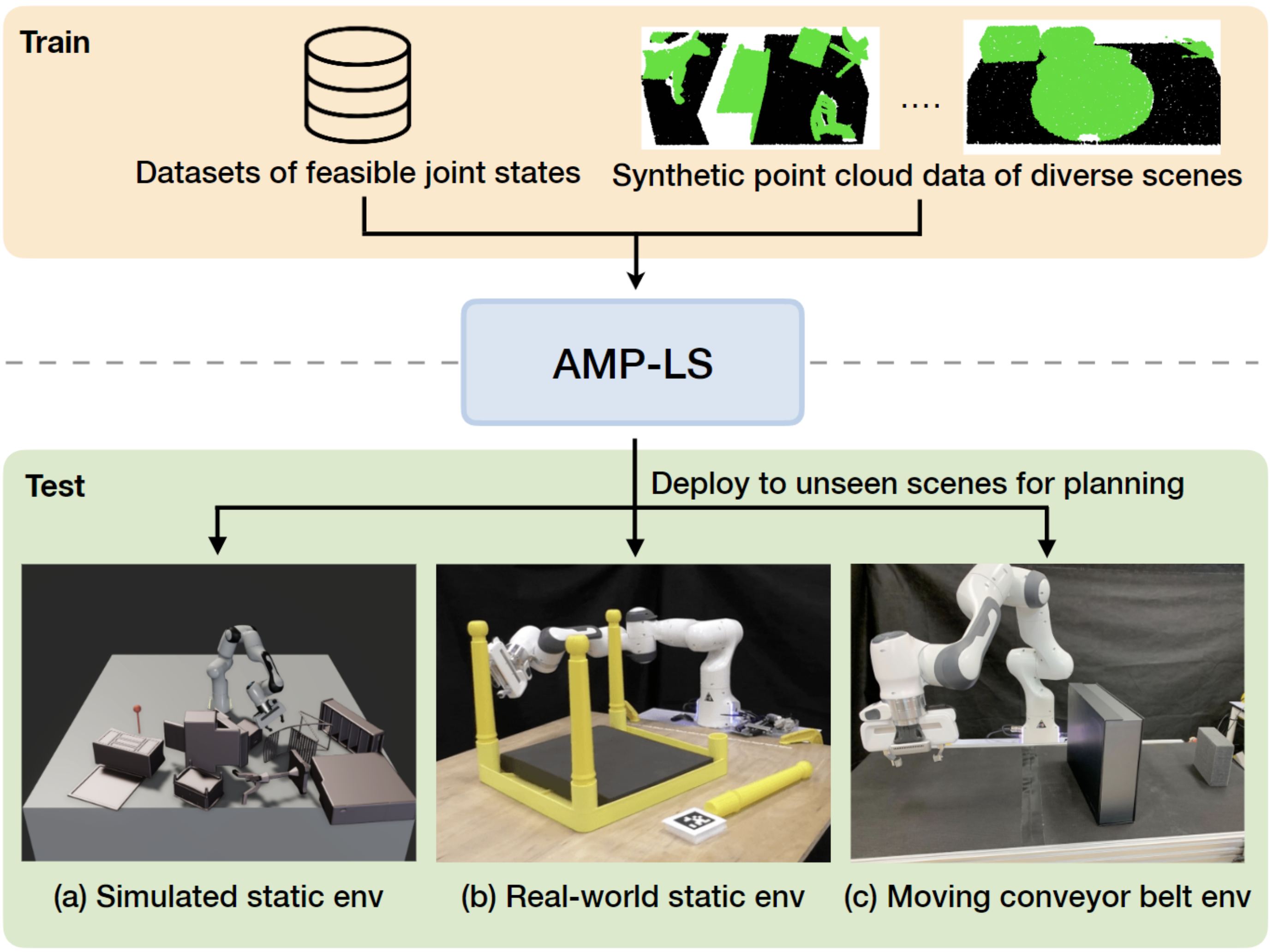}
    \caption{\textbf{Problem setup.} AMP-LS generates a collision-free trajectory via gradient-based optimisation by leveraging scene embeddings. Our model is trained on kinematically feasible robot joint states and synthetic point clouds of diverse scenes. For evaluation, our method is deployed to unseen scenes including: (a) \textit{Simulated static env}: Novel scenes generated by randomly placing obstacles on a table. (b) \textit{Real-world static env}: A robot avoids the table legs to reach the pre-grasp location of the unassembled table leg. (c) \textit{Moving Conveyor Belt env}: A robot reaches a moving target object while avoiding an obstacle on the conveyor belt by using closed-loop planning.} 
    \label{fig:teaser}
    \vspace{-1.2em}
\end{figure}

Recently, learning-based motion planning~\cite{qureshi2019motion, qureshi2020motion} has gained the attention of the robotics community with the promise of increased computational efficiency and faster planning speed.
Notably, Latent Space Path Planning (LSPP)~\cite{chiaman2022reaching} introduces motion planning via gradient-based optimisation in the latent space of a VAE. The success rate of LSPP is commensurate with that of commonly used sampling and gradient-based motion planners, but with significantly reduced planning time. 
By learning a structured latent space using kinematically feasible and easily generated robot states, a learned latent space that is optimised via activation maximisation (AM)~\cite{erhan2009am} can produce diverse and adaptive behaviours~\cite{mitchell2020first}.
However, LSPP relies on state-based obstacle representation with predefined object shapes, which do not easily transfer to real-world environments.

To address the limitation of LSPP, we introduce a method significantly extending the prior work by incorporating a collision predictor that leverages scene embeddings and efficient collision checking, which regularises the optimisation during planning for safe collision avoidance. We name this new method Activation Maximisation Planning in Latent Space (AMP-LS).
Specifically, we adapt SceneCollisionNet~\cite{danielczuk2021object}, trained on diverse synthetic point clouds of scenes generated with objects from ShapeNet datasets~\cite{chang2015shapenet}, for our purpose to facilitate zero-shot transfer to unseen environments, including real-world scenes (see Fig.~\ref{fig:teaser}).
Due to the speed of our approach, we also show that our method can be applied to closed-loop settings where both the obstacles and goal pose are moving.

The contributions of our work are threefold: (1) we present Activation Maximisation Planning in Latent Space (AMP-LS), which significantly extends LSPP  by incorporating a collision predictor that leverages scene embeddings and explicit collision checking in order to regularise optimisation when planning for obstacle avoidance;
(2) we empirically demonstrate that our approach can be zero-shot transferred to unseen scenes, including real-world environments, through the use of a collision predictor that is trained on diverse synthetic scenes;
(3) we show that our method can be applied to closed-loop settings with reactive behaviour, capable of reaching a \emph{moving} target while also avoiding a \emph{moving} obstacle.

\begin{figure*}
    \centering
    \includegraphics[width=0.95\textwidth]{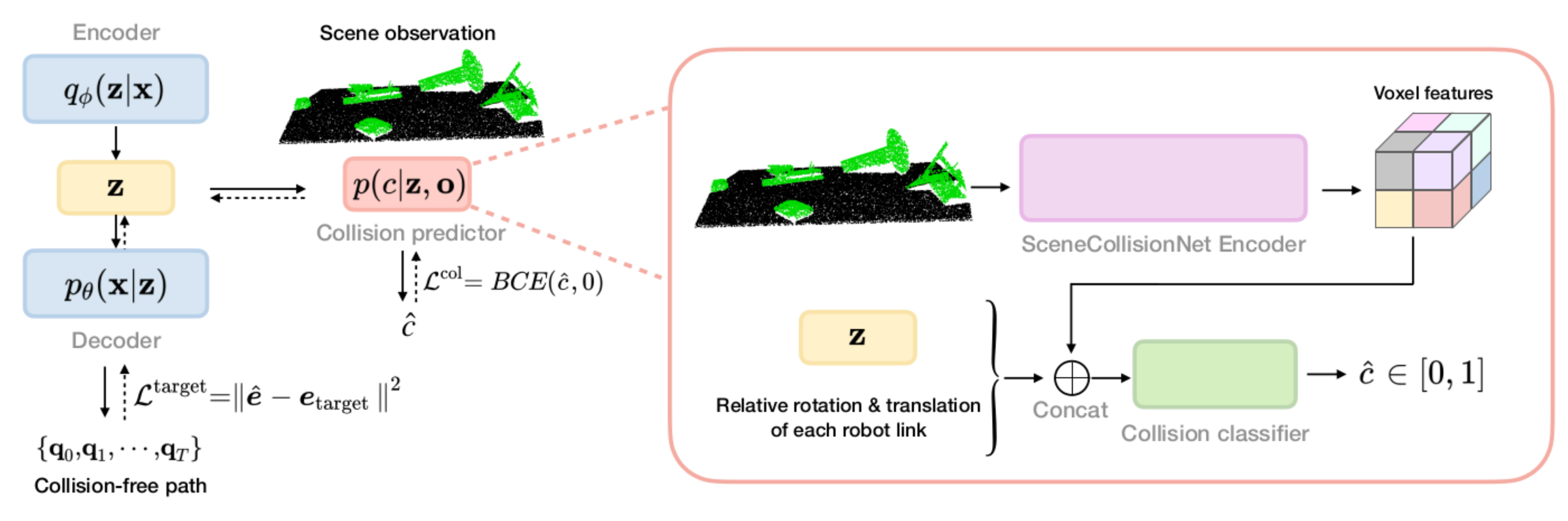}
    \caption{\textbf{Our method overview.} A VAE (\textcolor{blue}{\textbf{blue}}) is trained using feasible robot states $\mathbf{x}$ consisting of joint states, end-effector position, and end-effector orientation to learn structured latent representations $\mathbf{z}$ (\textcolor{yellow}{\textbf{yellow}}). Then, freezing the weights of the pre-trained encoder in the VAE, the collision predictor (\textcolor{red}{\textbf{red}}) takes as input the learned latent representation $\mathbf{z}$ and a scene point cloud observation $\mathbf{o}$. The collision predictor built upon SceneCollisionNet learns to output a probability $\hat{c}$ of collision between the robot arm and obstacles. To plan a collision-free trajectory, gradient-based optimisation is applied to produce a sequence of latent representations $\{\mathbf{z}_{t}\}_{t=1}^{T}$ each of which has a low probability of collision with the scene using the learned collision predictor. A sequence of joint states $\{\mathbf{q}_{t}\}_{t=1}^{T}$ is generated by decoding the sequence of latent representations $\{\mathbf{z}_{t}\}_{t=1}^{T}$ using the trained decoder in the VAE.}
    \label{fig:method}
    \vspace{-1.2em}
\end{figure*}

\section{Related Works}
Sampling-based motion planning approaches such as RRT~\cite{lavalle1998RRT, karaman2011RRTStar} and PRM~\cite{amato1996PRM} are widely used to generate collision-free trajectories in robotics.
PRM requires a pre-computed roadmap; RRT often struggles to find the solution with the shortest path. While several extensions such as RRT*~\cite{karaman2011RRTStar} and BIT*~\cite{gammell2020batch} have been proposed to achieve asymptotic optimality and reduce computational cost, these approaches typically demand many samples---a runtime problem that compounds with increases in robot DoF, environmental complexity, or path length~\cite{hauser2015lazy}.
Another limitation of sampling-based motion planners is that they do not support real-time planning, as re-planning is required to navigate dynamic environments.

Optimisation-based planning approaches such as covariant Hamiltonian optimisation for motion planning (CHOMP)~\cite{chomp} and Stochastic Trajectory Optimisation for Motion Planning (STOMP)~\cite{stomp} require a large number of trajectories when given multiple constraints. These approaches typically start from an initial guess, a trajectory linking the start and desired end states, which is refined through minimisation of a cost function. Computation terminates when a stop condition is met or the algorithm times out. 
The artificial potential algorithm \cite{khatib1985real, flacco2012depth} is perhaps the closest optimisation-based planning approach to our work. It achieves real-time obstacle avoidance by creating attractive and repulsive fields around goals and obstacles. End-effector movement is then guided by the gradient of these fields. Although appealing in its simplicity, it struggles to handle additional constraints on properties that cannot be fully determined by robot joint configuration.

Several recent works attempt to leverage neural networks for motion planning. 
Neural motion planning methods \cite{Pfeiffer_2017, ichter2018learning, qureshi2019motion, qureshi2018deeply} employ imitation learning (IL) on expert demonstrations generated by a sampling-based motion planner or reinforcement learning (RL)~\cite{Sutton1998} to learn motion policies. 
Notably, Motion Policy Network~\cite{fishman2022motion} achieves commensurate success rates when compared against traditional planning approaches and even generalises well to unseen environments. However, these methods require a large number of trajectories, often generated by an expert planner, to train a motion planning policy.

Another set of works performs planning in a learned latent space~\cite{Ichter2019robot, chiaman2022reaching}.
L2RRT~\cite{ichter2018learning} plans a path in a learned latent space using RRT. 
Our work builds upon \textit{Latent Space Path Planning} (LSPP)~\cite{chiaman2022reaching}. LSPP plans a trajectory for a robot via iterative optimisation using activation maximisation (AM)~\cite{erhan2009am} in a latent space of the robot kinematics learned by a generative model.
Leveraging a collision predictor as a constraint, LSPP successfully plans a collision-free path with improved efficiency in planning time.
However, LSPP approximates a scene as a set of cylindrical obstacles and requires state-based knowledge of the scene, such as position and shape of obstacles. Such narrow scene definitions and lack of complete information limits the application of this method to real-world problems.

To successfully generate a collision-free path in a scene with obstacles, learning a collision predictor to identify the collision between a robot and the scene is essential.
Prior neural motion planning methods  \cite{Pfeiffer_2017, ichter2018learning, qureshi2019motion,strudel2020learning} learn obstacle representations either from 2D images, occupancy grids, or point clouds, instead of explicitly predicting a probability of collision.
SceneCollisionNet~\cite{danielczuk2021object} learns the scene embeddings for a collision predictor from a large number of synthetic scenes generated with diverse objects from ShapeNet~\cite{chang2015shapenet}.
To leverage a collision predictor as a constraint for motion planning, we utilise SceneCollisionNet and adapt it to work within our latent planning framework.

\section{Approach}
In this work, we introduce a method significantly extending the prior work~\cite{chiaman2022reaching} and name it Activation Maximisation Planning in Latent Space (AMP-LS). 
Similar to the prior work~\cite{chiaman2022reaching}, AMP-LS leverages a variational autoencoder (VAE)~\cite{kingma2013auto, rezende2014stochastic} to learn a structured latent space to generate kinematically feasible joint trajectories.
While a collision predictor in the prior work relies on state-based obstacle representations, our collision predictor leverages scene embeddings obtained from SceneCollisionNet~\cite{danielczuk2021object} to readily achieve zero-shot transfer to unseen environments.
Further, we present an approach for explicit collision checking to directly regularise the optimisation to plan collision-free trajectories.
In the following section, we describe an overview of our model (see Fig.~\ref{fig:method}) and optimisation objective for planning.

\subsection{Problem Formulation}
Similar to LSPP~\cite{chiaman2022reaching}, we consider the problem of generating a collision-free trajectory consisting of robot joint configurations $\{\mathbf{q}_{0}, \dots, \mathbf{q}_{T}\}$ for a robot in an environment with obstacles.
A state $\mathbf{x}_{t}$ at time $t$ consists of a kinematically feasible robot joint configuration $\mathbf{q}_{t}$, and its end-effector position $\mathbf{e}^{pos}_{t}$ and orientation $\mathbf{e}^{ori}_{t}$.
While LSPP considers only the joint states and end-effector position, adding orientation is essential to tackle motion planning problems.
The end-effector orientation $\mathbf{e}^{ori}$ employs a 6D representation of SO(3), which consists of the first two column vectors in a rotation matrix $\mathbf{R}$. This representation is suitable for learning rotations using neural networks due to its property of continuity~\cite{Zhou_2019_CVPR}.
Note that no prior information of obstacles  (\eg mesh) is given.
In contrast to LSPP, which leverages low-dimensional state information as an observation, AMP-LS utilises point cloud observations $\mathbf{o}_{t} \in \mathcal{R}^{n\times 3}$ with $n$ points from a third-person camera, which includes only scene information. Thus, the robot point cloud is filtered out from the raw point cloud.

\subsection{Learning Latent Representations of Robot State}
To plan cohesive paths for the manipulator using a learned latent space, the latent space must be structured such that representations of similar joint states are close to each other.
Leveraging a VAE~\cite{kingma2013auto, rezende2014stochastic}, prior work~\cite{chiaman2022reaching} successfully learns such a latent space and captures a notion of local distance in joint space.
In their representation, poses that are close to each other in joint space are also close in latent space.
Similarly, we also learn a VAE consisting of an encoder $q_{\phi}(\mathbf{z}|\mathbf{x})$ and decoder $p_{\theta}(\mathbf{x}|\mathbf{z})$, where $\mathbf{z}$ is the latent representation.
To train the VAE, rather than directly maximise the evidence, $p_{\theta}(\mathbf{x})=\int p_{\theta}(\mathbf{x} \mid \mathbf{z}) p_{\theta}(\mathbf{z}) d \mathbf{z}$, which is generally intractable, we instead optimise the evidence lower bound (ELBO) $\mathcal{L}^{\mathrm{ELBO}} \leq p(\mathbf{x})$:

\begin{equation}
    \mathcal{L}^{\mathrm{ELBO}}=\underbrace{\mathbb{E}_{\mathbf{z} \sim q_{\phi}(\mathbf{z} \mid \mathbf{x})} \log p_{\theta}(\mathbf{x} \mid \mathbf{z})}_{\text {Reconstruction Accuracy }}-\underbrace{D_{\mathrm{KL}}\left[q_{\phi}(\mathbf{z} \mid \mathbf{x}) \| p(\mathbf{z})\right]}_{\text {KL Term }}
\end{equation}
There is a trade-off in the ELBO loss between the reconstruction accuracy and the KL term: accurate reconstruction at the cost of poorly structured latent space, on one hand, or well-structured latent space but noisy reconstruction, on the other.
These terms are often manually weighted in the ELBO formulation~\cite{higgins2017betavae}.
An alternative to manually tuning the weight is to use GECO~\cite{rezende2018taming}. GECO adaptively tunes the trade-off between reconstruction and regularisation by formulating the ELBO loss as a constrained optimisation problem with a Lagrange multiplier $\lambda$:

\begin{equation}
    \mathcal{L}^{\mathrm{GECO}}=\underbrace{-D_{\mathrm{KL}}\left[q_{\phi}(\mathbf{z} \mid \mathbf{x}) \| p(\mathbf{z})\right]}_{\mathrm{KL} \text { Term }}+\lambda \underbrace{\mathbb{E}_{\mathbf{z} \sim q_{\phi}(\mathbf{z} \mid \mathbf{x})}[\mathcal{C}(\mathbf{x}, \hat{\mathbf{x}})]}_{\text {Reconstruction Error Constraint }}
\end{equation}
This encourages the model to optimise the reconstruction accuracy first, until it reaches a predefined target. The KL term is then optimised. 
The generative model is trained on a dataset of kinematically feasible joint states of the robot.

\subsection{Activation Maximisation for Motion Planning}
Our goal is to plan a trajectory consisting of robot joint configurations towards a target pose.
That is, given a target end-effector position $\mathbf{e}^{pos}_{\text{target}}$ and orientation $\mathbf{e}^{ori}_{\text{target}}$, we expect our method to generate a sequence of joint configurations $\{\mathbf{q}_{0}, \dots, \mathbf{q}_{T}\}$ that leads a robot to the target pose.
Leveraging the trained VAE inspired by prior work~\cite{chiaman2022reaching}, we can compute such a sequence of robot joints by decoding the latent representation of the VAE model  $\{\mathbf{z}_{0}, \dots, \mathbf{z}_{T}\}$.
This sequence of the latent representation is computed in a probabilistic model through activation maximisation (AM)~\cite{erhan2009am}:
\begin{equation}
    \mathbf{z}_{t+1} = \mathbf{z}_{t} - \alpha_{\mathrm{AM}}\nabla \mathcal{L}^{\mathrm{AM}}
\end{equation}
where
\begin{equation}
    \begin{split}
        \mathcal{L}^{\mathrm{AM}}=\lambda_{\text {target}}(\underbrace{\left\|\hat{\boldsymbol{e}}^{pos} - \boldsymbol{e}^{pos}_{\text {target }}\right\|_{2}}_{\text{Target Position Loss}} + \underbrace{\left\|\hat{\boldsymbol{e}}^{ori} - \boldsymbol{e}^{ori}_{\text {target }}\right\|_{2}}_{\text {Target Orientation Loss }}) \\ 
        + \underbrace{(-\log p(\mathbf{z}))}_{\text{Prior Loss}}
    \end{split}
    \label{eq:base_am}
\end{equation}
In contrast to the prior work~\cite{chiaman2022reaching}, we also introduce an end-effector orientation constraint, which is generally useful for reaching a pre-grasp pose.
The first latent representation $\mathbf{z_{0}}$ is acquired by encoding the current/starting robot state $\mathbf{z}_{0} \sim q_{\phi}(\mathbf{z} | \mathbf{x}=\mathbf{x}_{0})$.
Note that model parameters are not updated, but only the parameterised latent variable $\mathbf{z}$ is iteratively updated.
The first two terms in $\mathcal{L}^{\mathrm{AM}}$ (Eq. \ref{eq:base_am}) guide the latent representation to decode robot joint states that approach the target pose.
The third term is the likelihood of the current representation under its prior, which is introduced in~\cite{chiaman2022reaching} to encourage the latent representation to stay close to the training distribution, thus decoding to kinematically feasible pair of joint position and end-effector pose.

\begin{algorithm}[t]
    \caption{Planning a collision-free path in latent space via activation maximisation}
    \label{alg:planning}
    \begin{algorithmic}[1]
    \renewcommand{\COMMENT}[1]{{\algorithmiccomment{#1}}}
    \STATE Initialise a buffer $D=\{\mathbf{q}_{0}\}$, $\lambda_{\text{pos}}$, $\lambda_{\text{ori}}$, $\lambda_{\text{col}}$, $\mathbf{q}_{\text{prev}} = \mathbf{q}_{0}$
    \STATE $\mathbf{z}_{0} \sim q_{\phi}(\mathbf{z}|\mathbf{x}=\mathbf{x}_{0})$
    \FOR{$t=0, 1, 2, \dots, H$}
        \STATE $\{\hat{\mathbf{q}}_{t}, \hat{\mathbf{e}}^{pos}_{t}, \hat{\mathbf{e}}^{ori}_{t}\} \sim p_{\theta}(\mathbf{x}|\mathbf{z}=\mathbf{z}_{t})$
        \IF{$t > 0$ and $p_{\vartheta}(\mathbf{z}_{t}, \mathbf{o}_{t}) < \gamma_{\text{col}}$}
            \STATE $\{\mathbf{q}_{\text{prev}},\dots, \hat{\mathbf{q}}_{t}\} = f_{\text{interpolate}}(\mathbf{q}_{\text{prev}}, \hat{\mathbf{q}}_{t})$ \\
            \COMMENT{Linear interpolation between $\mathbf{q}_{\text{prev}}$ and $\hat{\mathbf{q}}_{t}$}
            \IF{collision in $\{\mathbf{q}_{\text{prev}},\dots, \hat{\mathbf{q}}_{t}\}$}
                \STATE $i \leftarrow$ index of the first joint state with collision in the interpolated trajectory
                \STATE $m \leftarrow |\{\mathbf{q}_{\text{prev}},\dots, \hat{\mathbf{q}}_{t}\}|$ \\
                \STATE Reduce $\lambda_{\text{pos}}$ and $\lambda_{\text{ori}}$ by a factor of $\frac{i}{m}$\\
                \STATE $\hat{\mathbf{q}}_{t} \leftarrow \mathbf{q}_{\text{prev}}$, $\mathbf{z}_{t} \leftarrow \mathbf{z}_{\text{prev}}$ \\
                \COMMENT{Back trace to the previous joint and latent representations $\mathbf{q}_{\text{prev}}$ and $\mathbf{z}_{\text{prev}}$ for replanning}
            \ELSE
                \STATE $D \leftarrow D \cup \{\mathbf{q}_{\text{prev}}, \dots \hat{\mathbf{q}}_{t}\}$
                \IF{$d(\hat{\boldsymbol{e}}_{t}, \boldsymbol{e}_{\text{target}}) < \gamma$}
                    \STATE break
                \ENDIF
                \STATE $\mathbf{q}_{\text{prev}} \leftarrow \hat{\mathbf{q}}_{t}$, $\mathbf{z}_{\text{prev}} \leftarrow \mathbf{z}_{t}$
            \ENDIF
        \ENDIF
        \STATE Compute losses (Eq. \ref{eq:am_collision})
        \STATE Update $\lambda_{\text{pos}}$, $\lambda_{\text{ori}}$, and $\lambda_{\text{col}}$ using GECO
        \STATE $\mathbf{z}_{t+1} \leftarrow \mathbf{z}_{t} - \alpha_{\mathrm{AM}} \nabla \mathcal{L}_{t}^{\mathrm{AM}}$ \\
    \ENDFOR
    \end{algorithmic}
\end{algorithm}

\subsection{Collision Constraints}
To generate a collision-free trajectory, similar to that used in prior work~\cite{chiaman2022reaching}, we add collision constraints to the objective function in Eq.~\ref{eq:base_am} by introducing a collision predictor.
While the prior work uses narrowly defined state-based obstacle representations as input to the collision predictor, in our approach, we adapt SceneCollisionNet~\cite{danielczuk2021object} to embed scene observations
for zero-shot transfer to unseen environments.
The voxel features from SceneCollisionNet are concatenated with the latent representation of the VAE $\mathbf{z}$ and the rotation and relative translation from each robot link to the centre of the closest voxel to form the input to the collision classifier.
The classifier predicts the probability of collision $\hat{c}$ between the robot and obstacles (see Fig.~\ref{fig:method}).
Note that we train the collision predictor only on features of voxels closest to each robot link to ignore unnecessary voxel information.
While training the collision predictor, the weights of the pre-trained VAE are frozen so that the pre-trained latent space does not change. The collision predictor is trained using the binary cross-entropy (BCE) loss with ground truth collision labels.
To drive the latent representation away from obstacles, we incorporate the collision predictor loss into Eq.~\ref{eq:base_am}:

\begin{equation}
\begin{split}
    \mathcal{L}^{\mathrm{AM}}=\lambda_{\text {target}}(\underbrace{\left\|\hat{\boldsymbol{e}}^{pos} - \boldsymbol{e}^{pos}_{\text {target }}\right\|_{2}}_{\text{Target Position Loss}} + \underbrace{\left\|\hat{\boldsymbol{e}}^{ori} - \boldsymbol{e}^{ori}_{\text {target }}\right\|_{2}}_{\text {Target Orientation Loss }}) \\ + \lambda_{\text {col}} \underbrace{\left(-\log \left(1-p_{\vartheta}\left(\mathbf{z}, \mathbf{o}\right)\right)\right)}_{\text {Collision Loss }} + \underbrace{(-\log p(\mathbf{z}))}_{\text{Prior Loss}}
    \label{eq:am_collision}
    \end{split}
\end{equation}
During planning, three coefficients $\lambda_{\text{target}}$ and $\lambda_{\text{col}}$ are automatically and dynamically adjusted by GECO~\cite{rezende2018taming}.
Minimising the collision loss during AM optimisation drives the latent representation $\mathbf{z}$ towards the representation whose decoded joint configuration is collision-free.

\subsection{Collision Checking}
While prior work~\cite{chiaman2022reaching} simply optimises the objective function until it reaches a target, we observe that it is hard to perfectly balance multiple loss terms and that such simple optimisation often results in collision between the robot and obstacles.
In contrast to the target losses, the collision loss is inherently a hard constraint that should not be violated at any point in the trajectory.
To address this issue, our high-level idea is that collision can be predicted and avoided before execution and the coefficients of the objective function determine the direction in which the latent representation is heading towards.
Specifically, we introduce explicit collision checking using the learned collision predictor and automatic rescaling for coefficients during the planning to avoid obstacles more safely.
That is, if a collision probability of the decoded joint configuration is higher than a predefined threshold $\gamma_{\text{col}}$, we reject such robot configuration that is highly likely to be in collision and keep optimising the latent space until the decoded joint state is collision-free.
Then, we interpolate a trajectory between the current and decoded collision-free joint state in $m$ steps and pass them to the collision predictor to check for collision.
If there is any collision in the interpolated trajectory, we obtain its index $i$ of the joint state with collision closest to the current joint state and reduce the coefficient of the target position and orientation loss by multiplying by $\frac{i}{m}$, to encourage the optimisation to minimise the collision loss.
Intuitively, this scaling induces the robot to deviate from the original route drastically depending on how close it is to an obstacle.
This process continues until the collision-free next joint state is found and there is no collision in the interpolated trajectory between the current joint state and the next joint state.
In our experiments, we use the threshold of $\gamma_{\text{col}} = 0.4$.
For further details, see Algorithm~\ref{alg:planning}.

\section{Implementation Details}

\begin{figure*}[t!]
    \centering
    \includegraphics[width=1\textwidth]{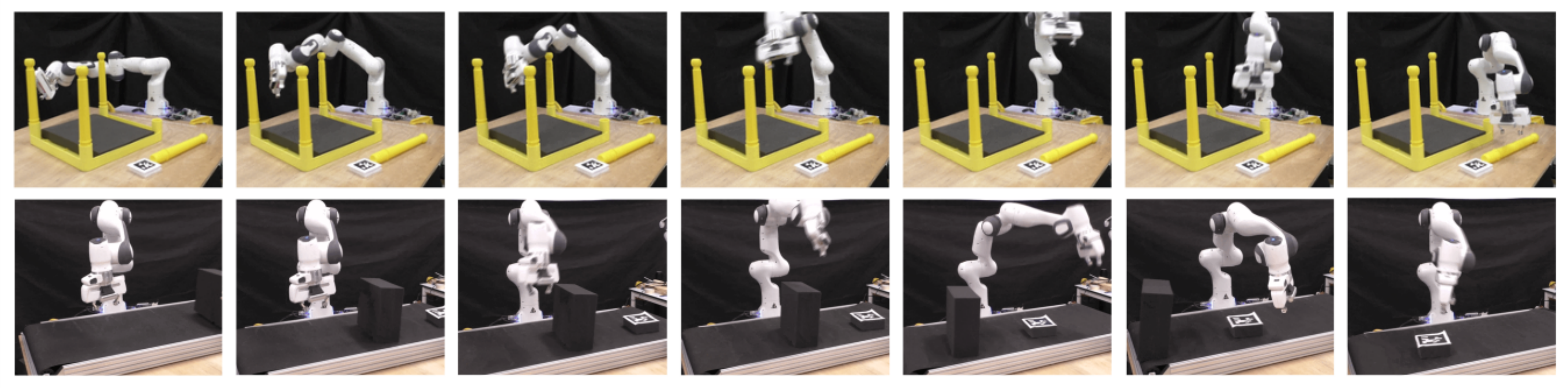}
    \caption{Visualisation of real-world experiments. \textbf{Top}: Our method successfully plans a collision-free trajectory in a complex real-world scene from an impeded start configuration to a pre-grasp goal configuration. By training a collision predictor on diverse synthetic scenes, our method can readily transfer to such unseen scene. \textbf{Bottom}: AMP-LS can be applied to closed-loop planning to avoid moving obstacles and reach a moving target object on a conveyor. This reader is referred to our supplementary video for better visualisation.}
    \label{fig:real_world_exp}
    \vspace{-1.2em}
\end{figure*}

\subsection{Architecture Details}
Our VAE encoder and decoder  consist of three fully connected hidden layers with $512$ units and ELU activation functions~\cite{clevert2015fast}. 
The input dimension to the VAE is $16$, consisting of robot joint states $\mathbf{q} \in \mathcal{R}^{7}$, end-effector position $\mathbf{e}^{pos} \in \mathcal{R}^{3}$ and 6D representation of end-effector rotation matrix $\mathbf{e}^{ori} \in \mathcal{R}^{6}.$
The dimension of the latent space $\mathbf{z}$ is $7$.
The collision classifier consists of fully connected layers with units of $[1024, 256]$.

\subsection{Training Details}
The VAE is trained using kinematically feasible robot joint configurations. 
To generate such joint states, we leverage the Flexible Collision Library (FCL)~\cite{pan2012fcl} for self-collision checking.
The VAE model is trained with a batch size of $256$ for about $2M$ training iterations using the Adam optimiser~\cite{kingma2015iclr} with a learning rate of $3\mathrm{e}{-4}$ on a GeForce RTX 3090. 
Throughout the training, valid robot configurations are generated on the fly as it is cheap to do so. In total, the model is exposed to around $500M$ configurations.

The collision predictor is trained on diverse synthetic point cloud data to assist zero-shot transfer to scenes with unseen obstacles.
Such scenes are generated by placing objects randomly sampled from the ShapeNet dataset~\cite{chang2015shapenet}, consisting of $8828$ 3D meshes.
Each object is placed on a planar surface with a random position and rotation.
We sample the number of objects placed on the surface from a uniform distribution between $4$ and $8$.
To train the collision predictor, a new scene is procedurally generated for each training iteration similar to the prior work~\cite{danielczuk2021object}, and we randomly sample $2048$ instances of kinematically feasible robot joint configurations and check for collisions between each robot configuration and the generated scene using FCL.
A third-person RGB-D camera is directed towards the centre of the scene to sample point clouds. The camera extrinsics  are randomly sampled for each query from a predefined range of roll, yaw, and pitch parameters.
Thus, $2048$ unique valid robot configurations and point clouds are procedurally generated for each iteration to train the collision predictor.
We train the collision predictor for $1M$ training iterations using SGD with a learning rate of $1\mathrm{e}{-3}$ and with momentum $0.9$ for approximately $7$ days, which is similar to the training time requirement of SceneCollisionNet.

\subsection{Deploymenet details}
In open-loop planning, the current state $\textbf{x}_{0}$ is encoded to a latent representation $\textbf{z}_{0}$.
Then, the encoded latent representation is iteratively optimised through AM optimisation (see Eq. \ref{eq:am_collision}) until the end-effector reaches the target pose with a tolerance of $\gamma$.
In closed-loop planning, while the latent representation is similarly optimised, a point cloud input for the collision predictor and the target pose in the objective function (see Eq.~\ref{eq:am_collision}) are updated at each time step for reactive motion.

\section{Experiments}
We design our experiments to answer the following guiding questions: (1) how does AMP-LS perform compared to traditional motion planning methods such as sampling and optimisation-based approaches in open-loop settings? (2) does AMP-LS transfer zero-shot to real-world static environments? (3) does AMP-LS cope with dynamic environments using closed-loop planning?

\subsection{Experimental Setup}
We evaluate our approach in both simulated and real-world environments.
In simulated experiments, we use the Gazebo simulator~\cite{koenig2004gazebo} with ROS.
In all of the simulated and real-world experiments, we use a 7-DoF Franka Panda robot. 

\subsection{Open-Loop Planning for Reaching Static Targets}
We evaluate AMP-LS in an open-loop planning setup in a simulated environment.
In this experiment, obstacles in the environment are static.
We select a range of sampling and optimisation-based motion planners typically used by the robotics community and available within the unified MoveIt! library. We compare our method against several sampling-based motion planners and an optimisation-based motion planner: RRT-Connect~\cite{kuffner2000rrt_connect}, RRT*~\cite{karaman2011RRTStar}, Lazy PRM*~\cite{bohlin2000larzyPRM}, LBKPIECE~\cite{csucan2009kinodynamic}, BIT*~\cite{gammell2020batch}, and CHOMP~\cite{chomp}.
CHOMP uses a linear initialisation from start to goal joint positions.
Since we assume that complete knowledge of the environment is not available, occupancy maps~\cite{hornung13auro} generated from point clouds are used for collision checking in motion planning baseline methods.
We evaluate the methods on $100$ novel scenes where objects are randomly placed on a table (see Fig.~\ref{fig:teaser} (a)).
The hyperparameters used for GECO to determine coefficients of our objective function (see Eq. \ref{eq:am_collision}) are found via a grid search similar to that of prior work~\cite{chiaman2022reaching}.
For the baselines, we use the default parameters provided by MoveIt OMPL. For RRT*, Lazy PRM*, and BIT*, the same $1$ second planning budget is given.
Across all methods, a motion plan is considered to be successful if a robot reaches a target within a distance tolerance of $1$cm and orientation tolerance of $15$ degrees.

As illustrated in Table~\ref{tab:main_result}, our method achieves a reasonable success rate with improved planning time compared to most of the motion planning baselines.
Specifically, AMP-LS outperforms CHOMP, which is also an optimisation-based motion planner, by a significant margin because CHOMP requires a large number of trajectories to find a feasible path in complex scenes, in contrast to AMP-LS.
AMP-LS still has a commensurate success rate against RRT-Connect, but the planning time of AMP-LS is an order of magnitude faster than the baseline.
Traditional motion planning baselines often fail to find a collision-free path within a short time and sometimes plan a path with collision due to occlusions in the scenes.
In contrast, our collision predictor is trained on diverse synthetic scenes with occlusion and can therefore reason about occluded regions, similar to SceneCollisionNet~\cite{danielczuk2021object}.
While our method demonstrates reasonable accuracy and improved planning efficiency, the path length is longer than most of the other baselines.
The longer path length is due to the design of the planning strategy~\cite{chiaman2022reaching} that tunes the coefficients of losses automatically to avoid obstacles, thus not directly minimising the path length.
To address this issue, additional optimisation constraints could be explored in the future that focus on reducing the path length.

As illustrated in Table~\ref{tab:main_result}, we also ablate constraints, such as prior loss, collision loss, and explicit collision checking.
The success rate of AMP-LS without a collision loss significantly drops, indicating that our collision predictor successfully constrains the latent space even in novel scenes.
Similar to the prior work~\cite{chiaman2022reaching}, AMP-LS without the prior loss results in significantly poorer performance as the latent representation is optimised to drive into unseen latent representations, which decode to kinematically inconsistent configurations.
Furthermore, Table~\ref{tab:main_result} shows that the success rate of AMP-LS without explicit collision checking drastically decreases because perfectly optimising multiple loss terms is often challenging, resulting in the collision with obstacles.

\begin{table}
    \centering
    \resizebox{\linewidth}{!}{%
    \begin{tabular}{l||c c c}
          &  Success rate & Planning time (s) & Path length \\
         \hline
          AMP-LS (ours) & \textbf{0.88 $\pm$ 0.06} & \textbf{0.16 $\pm$ 0.13} & 3.61 $\pm$ 1.05 \\
         \hline
         AMP-LS w/o col. loss & 0.46 $\pm$ 0.10 & 0.12 $\pm$ 0.04 & 3.68 $\pm$ 1.29 \\
         AMP-LS w/o prior loss & 0.35 $\pm$ 0.09 & 0.24 $\pm$ 0.21 & 3.23 $\pm$ 1.12 \\
         AMP-LS w/o explicit collision & 0.75 $\pm$ 0.08 & 0.15 $\pm$ 0.21 & 3.50 $\pm$ 1.12 \\
         \hline
         RRT-Connect & \textbf{0.86 $\pm$ 0.07} & 1.60 $\pm$ 0.89  & \textbf{2.17 $\pm$ 0.84} \\
         RRT* & 0.36 $\pm$ 0.09 & N/A & 2.25 $\pm$ 0.78 \\
         Lazy PRM* & \textbf{0.82 $\pm$ 0.08} & N/A & 2.26 $\pm$ 0.82 \\
         LBKPIECE & 0.23 $\pm$ 0.08 &  2.54 $\pm$ 1.12 & 2.34 $\pm$ 0.92 \\
         BIT* & 0.63 $\pm$ 0.09 & N/A & 2.42 $\pm$ 1.02 \\
         CHOMP & 0.39 $\pm$ 0.10 & 2.24 $\pm$ 0.79 & 2.41 $\pm$ 0.90
    \end{tabular}}
    \caption{Comparison of performance of our method AMP-LS against baseline motion planning algorithms with ablations. We also report $95\%$ confidence interval of Wilson score~\cite{wilson1927} for success rate and standard deviation for planning time and path length. The path length is normalised by dividing the actual path length by the distance between the initial and target end-effector positions for fairer comparison.}
    \label{tab:main_result}
\end{table}

\begin{figure}[h]
    \centering
    \includegraphics[width=0.43\textwidth]{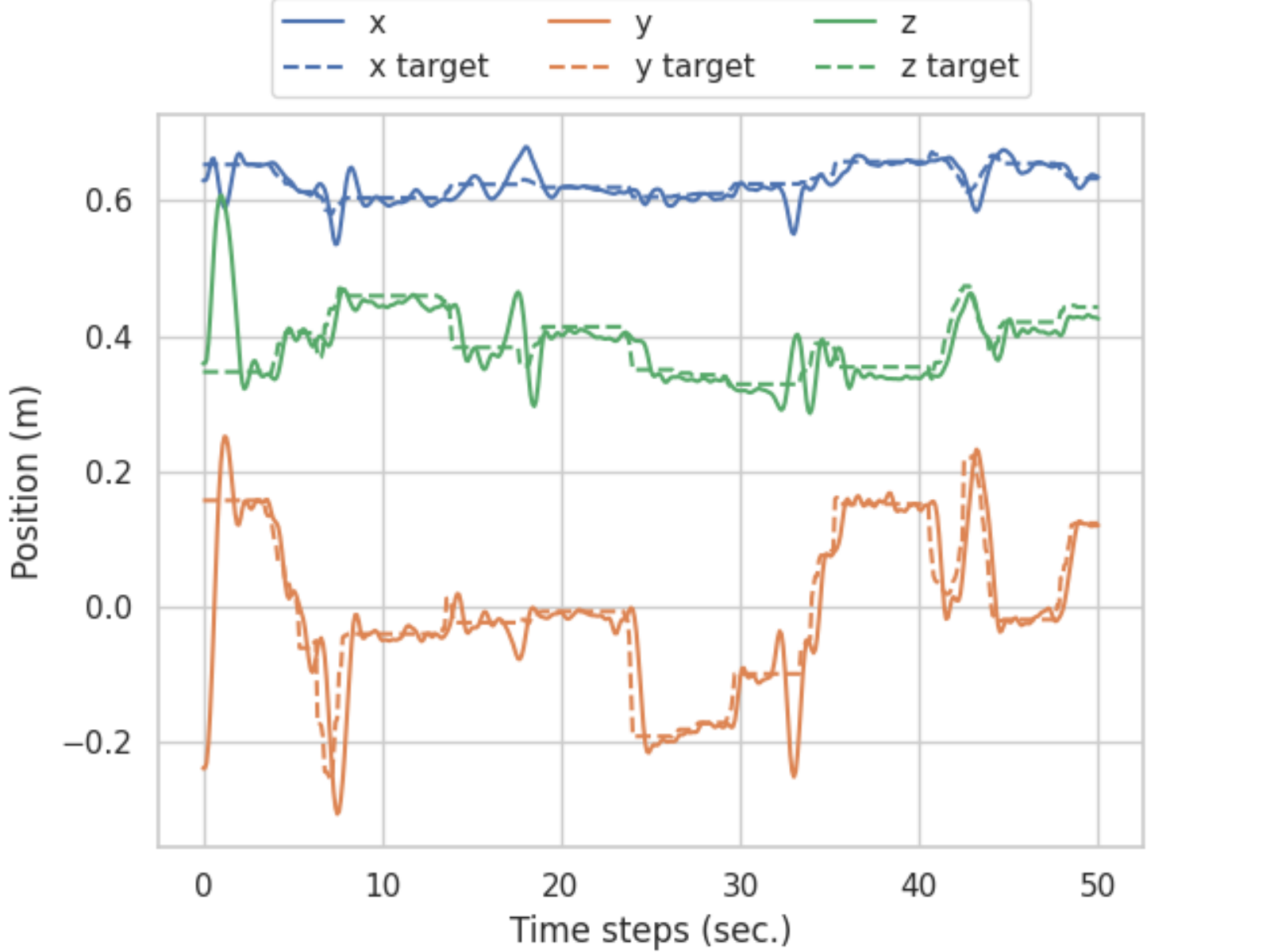}
    \caption{\textbf{Coordinates of end-effector and moving targets in closed-loop settings.} To verify the ability of closed-loop planning in our method, we deploy our method to the real-world robot arm to reach a moving target. }
    \label{fig:closed_loop}
    \vspace{-1.1em}
\end{figure}

\subsection{Real-World Open-Loop Planning in a Complex Scene}
Our method readily transfers to complex real-world scenes. To verify this, we qualitatively evaluate our method in a complex real-world static scene using open-loop planning as illustrated in Fig.~\ref{fig:teaser} (c).
In this task, the robot needs to reach the unassembled table leg while avoiding the other table legs to achieve a pre-grasp pose in a furniture assembly task.
We control the robot arm using an impedance controller.
As shown in Fig.~\ref{fig:real_world_exp} Top, our method can successfully plan a collision-free trajectory for a robot starting next to the table legs to avoid obstacles and reach the unassembled table leg on the table.
This demonstrates that our collision predictor, trained on diverse synthetic scenes, is transferable to real-world environments.

\subsection{Closed-Loop Planning for Moving Obstacles and Targets}
As our method is, by design, an efficient local planner, AMP-LS is able to act reactively when operated as a closed-loop system.
To verify the closed-loop potential of AMP-LS, we deploy our method on a robot with the goal of reaching a moving target without obstacles.
To control the real-world robot, a desired next joint position is sent to an impedance controller at $10$Hz. 
Fig.~\ref{fig:closed_loop} illustrates coordinates of the moving target and the end-effector position over $50$ seconds.
Since our method can predict the next desired joint state quickly, the robot can reactively follow the moving target.

To further demonstrate the ability of reactive motion using AMP-LS, we evaluate our method on a setup where the robot needs to avoid moving obstacles and reach a target object on a conveyor in both simulated and real-world environments (see Fig.~\ref{fig:real_world_exp} Bottom).
Firstly, we quantitatively evaluate our method to examine the ability of reactive motion in the simulated environment.
In this evaluation, we randomly generate obstacles of different sizes, and the obstacle and a target object are randomly placed on the conveyor belt.
We observe that the robot successfully avoids the obstacle and reaches a moving target on the conveyor with a success rate of $93.3\%$ ($28/30$ trials) thanks to the fast planning of our method. Note that we use a threshold of $3$cm and $20$ degrees in this experiment, because tight tolerance for reaching a moving target is challenging unless a future state of the target is estimated and used for planning.

In the real-world experiment, the robot starts moving towards the target object with attached AprilTag~\cite{olson2011apriltag} that is tracked by the third-person camera. For closed-loop planning, the collision predictor takes as input a point cloud for each time step.
As illustrated in Fig.~\ref{fig:real_world_exp} Bottom, the robot successfully avoids the moving obstacle to reach and follow the target object.

\section{Conclusion}
In this work, we present AMP-LS, a learning-based motion planning approach that generalises to unseen obstacles in complex environments.
AMP-LS builds upon LSPP~\cite{chiaman2022reaching} and inherits a number of desirable properties.
However, AMP-LS considerably extends LSPP by introducing a collision predictor trained on diverse synthetic scenes to leverage scene embeddings for unseen scene generalisation, and explicit collision checking during planning for safe obstacle avoidance.
We demonstrate that AMP-LS successfully generates collision-free paths in both unseen simulated and real-world scenes.
The comparison between AMP-LS and several sampling and optimisation-based motion planning baselines shows that our method achieves a commensurate success rate with much improved planning time. 
Furthermore, our real-world experiments show that AMP-LS can handle both open and closed-loop planning, which significantly broadens the applicability to real-world robotic problems.

\section*{ACKNOWLEDGMENT}
This work was supported by a UKRI/EPSRC Programme Grant [EP/V000748/1], we would also like to thank the University of Oxford for providing Advanced Research Computing (ARC) facility in carrying out this work  (\url{http://dx.doi.org/10.5281/zenodo.22558}).

\addtolength{\textheight}{-3.5cm}   %

\clearpage
\bibliographystyle{IEEEtran}
\bibliography{bib/conference, bib/robotics, bib/env, bib/motion_planning, bib/deep_learning, bib/misc}

\end{document}